\documentclass[letterpaper]{article} 
\usepackage{aaai25}  
\usepackage{times}  
\usepackage{helvet}  
\usepackage{courier}  
\usepackage[hyphens]{url}  
\usepackage{graphicx} 
\usepackage{natbib}  
\usepackage{caption} 
\usepackage{algorithm}
\usepackage{algorithmic}
\usepackage{multirow}
\usepackage{booktabs}
\usepackage{amsmath}
\usepackage{amssymb}
\usepackage{subcaption}

\usepackage{newfloat}
\usepackage{listings}
\DeclareCaptionStyle{ruled}{labelfont=normalfont,labelsep=colon,strut=off} 
\floatstyle{ruled}
\newfloat{listing}{tb}{lst}{}
\floatname{listing}{Listing}
\title{Efficient Rectification of Neuro-Symbolic Reasoning Inconsistencies\\ by Abductive Reflection}
\author {
    Wen-Chao Hu\textsuperscript{\rm 1,2}, 
    Wang-Zhou Dai\textsuperscript{\rm 1,3}, 
    Yuan Jiang\textsuperscript{\rm 1,2}, 
    Zhi-Hua Zhou\textsuperscript{\rm 1,2}
}
\affiliations {
    \textsuperscript{\rm 1}National Key Laboratory for Novel Software Technology, Nanjing University, China\\
    \textsuperscript{\rm 2}School of Artificial Intelligence, Nanjing University, China\\
    \textsuperscript{\rm 3}School of Intelligence Science and Technology, Nanjing University, China\\
    \{huwc, daiwz, jiangy, zhouzh\}@lamda.nju.edu.cn
}

\begin{document}

\maketitle

\begin{abstract}
Neuro-Symbolic (NeSy) AI could be regarded as an analogy to human dual-process cognition, modeling the intuitive System 1 with neural networks and the algorithmic System 2 with symbolic reasoning. However, for complex learning targets, NeSy systems often generate outputs inconsistent with domain knowledge and it is challenging to rectify them. Inspired by the human Cognitive Reflection, which promptly detects errors in our intuitive response and revises them by invoking the System 2 reasoning, we propose to improve NeSy systems by introducing Abductive Reflection (ABL-Refl) based on the Abductive Learning (ABL) framework. ABL-Refl leverages domain knowledge to abduce a reflection vector during training, which can then flag potential errors in the neural network outputs and invoke abduction to rectify them and generate consistent outputs during inference. ABL-Refl is highly efficient in contrast to previous ABL implementations. Experiments show that ABL-Refl outperforms state-of-the-art NeSy methods, achieving excellent accuracy with fewer training resources and enhanced efficiency.
\end{abstract}

%

\section{Introduction}

Human decision-making is generally recognized as an interaction between two systems: System 1 quickly generates an intuitive response, and System 2 engages in further algorithmic and slow reasoning~\cite{frederick2005cognitive,kahneman2011thinking}. In Neuro-Symbolic (NeSy) Artificial Intelligence (AI), neural networks often resemble System 1 for rapid pattern recognition, and symbolic reasoning mirrors System 2 to leverage domain knowledge and handle complex problems thoughtfully, yet in a slower and more controlled way~\cite{bengio2019system}. Like human System 1 reasoning, when facing complicated tasks, neural networks often produce unreliable outputs which cause inconsistencies with domain knowledge. These inconsistencies can then be reconciled with the help of the symbolic reasoning counterpart~\cite{hitzler2022neuro}. 

To achieve the above process, some methods relax symbolic domain knowledge as neural network constraints~\cite{xu2018semantic,yang2022injecting}, some attempt to approximate logical calculus using distributed representations within neural networks~\cite{wang2019satnet}. However, a loss of full symbolic reasoning ability often occurs during these relaxation or approximation, hampering the ability of generating reliable output.

Abductive Learning (ABL)~\cite{zhou2019abductive,zhou2022abl} is a framework for bridging machine learning and logical reasoning while preserving full expressive power in each side. In ABL, the machine learning component first converts raw data into primitive symbolic outputs. These outputs can be utilized by the symbolic reasoning component, which leverages domain knowledge and performs abduction to generate a revised, more reliable output. However, previous implementations of ABL require a highly discrete combinatorial consistency optimization before applying abduction, and this optimization has high complexity which encumbers, thereby severely limiting the efficiency and applicability to large-scale scenarios.

Human reasoning naturally exploits both sides efficiently, a hypothetical model for this process is called Cognitive Reflection, where the fast System 1 thinking is called to quickly generate an approximate over-all solution, and then seamlessly hands the complicated parts to System 2~\cite{frederick2005cognitive}. The key to this process is the reflection mechanism, which promptly detects which part in the intuitive response may contain inconsistencies with domain knowledge and invokes System 2 to rectify them. This reflection typically positively associates with System 2 capabilities, as both are closely linked to an individual's mastery of domain knowledge~\cite{sinayev2015cognitive}. Following the reflection, the process of the step-by-step formal reasoning becomes less complex: With a largely reduced search space, deriving the correct solution for System 2 becomes straightforward.

Inspired by this phenomenon, we propose a general enhancement, \textit{Abductive Reflection (ABL-Refl)}. Based on ABL framework, ABL-Refl preserves full expressive power of neural networks and symbolic reasoning, while replacing the time-consuming consistency optimization with the reflection mechanism, thereby significantly improves efficiency and applicability. Specifically, in ABL-Refl, a reflection vector is concurrently generated with the neural network intuitive output, which flags potential errors in the output and invokes symbolic reasoning to perform abduction, thereby rectifying these errors and generating a new output that is more consistent with domain knowledge. During model training, the training information for the reflection derives from domain knowledge. In essence, the reflection vector is abduced from domain knowledge and serves as an attention mechanism for narrowing the problem space of symbolic reasoning. The reflection can be trained unsupervisedly, requiring only the same amount of domain knowledge as state-of-the-art NeSy systems without generating extra training data.

We validate the effectiveness of ABL-Refl in solving Sudoku NeSy benchmarks in both symbolic and visual forms. Compared to previous NeSy methods, ABL-Refl performs significantly better, achieving higher reasoning accuracy efficiently with fewer training resources. We also compare our method to symbolic solvers, and show that the reduced search space in ABL-Refl improves the reasoning efficiency. Further experiments on solving combinatorial optimization on graphs validate that ABL-Refl can handle diverse types of data in varied dimensions, and exploit knowledge base in different forms.

\section{Related Work}

Recently, there has been notable progress in enhancing neural networks with reliable symbolic reasoning. Some methods use differentiable fuzzy logic~\cite{serafini2016logic,marra2020integrating} or relax symbolic domain knowledge as constraints for neural network training~\cite{xu2018semantic,yang2022injecting,hoernle2022multiplexnet,ahmed2022neuro}, while others learn constraints within neural networks by approximating logic reasoning with distributed representations~\cite{amos2017optnet,selsam2018learning,wang2019satnet}. These models tend to soften the requirements in symbolic reasoning, impacting the reliability of output generation. Models like DeepProbLog~\cite{manhaeve2018deepproblog} and NeurASP~\cite{yang2020neurasp} interpret the neural network output as a distribution over symbols and then apply a symbolic solver, incurring substantial computational costs. Abductive Learning (ABL)~\cite{zhou2019abductive,zhou2022abl} attempts to integrate machine learning and logical reasoning in a balanced and mutually supporting way. It features an easy-to-use open-source toolkit~\cite{huang2024ablkit} with many practical applications~\cite{SS-ABL2020Huang,GABL2021Cai,Tac2021Wang,KESAR2024Gao}. However, the consistency optimization is with high complexity.

Another category of work related to our study also follows a similar process of prediction, error identification, and reasoning~\cite{nair2020solving,nye2021improving,han2023gnn}. These methods are usually constrained in a narrow scope of domain knowledge, confined to specific mathematical problems or are bounded within a minimal world model.

Cornelio \textit{el al.}~\shortcite{cornelio2023learning} generates a selection module to identify errors requiring symbolic reasoning rectification. In constrast to their approach which requires the preparation of a large synthetic dataset in advance, our approach automatically abduces the reflection vector during model training. 

\section{Abductive Reflection}

This section presents problem setting and the \textit{Abductive Reflection (ABL-Refl)} method.

\subsection{Problem Setting}

The main task of this paper is as follows: The input is raw data $\boldsymbol{x}$, which can be in either symbolic or sub-symbolic form, and the target output is $\boldsymbol{y}=\left[y_1,y_2,\dots,y_n\right]$, with each $y_i$ being a symbol from a set $\mathcal{Y}$ that contains all possible output symbols. We assume two key components at our disposal: neural network $f$ and domain knowledge base $\mathcal{KB}$. $f$ can directly map $\boldsymbol{x}$ to $\boldsymbol{y}$, and $\mathcal{KB}$ holds constraints between the symbols in $\boldsymbol{y}$. $\mathcal{KB}$ can assume various forms, including propositional logic, first-order logic, mathematical or physical equations, etc., and can perform symbolic reasoning operations by exploiting the corresponding symbolic solver. The output $\boldsymbol{y}$ should adhere to the constraints in $\mathcal{KB}$, otherwise it will inevitably contain errors that lead to inconsistencies with the domain knowledge and incorrect reasoning results.

This problem type has broad applications. For example, it can be used to solve Sudoku puzzles, where the output $\boldsymbol{{y}}$ consists of $n=81$ symbols from the set $\mathcal{Y}=\{1,2,\dots,9\}$, and the constraints in $\mathcal{KB}$ are the rules of Sudoku. It can also be applied in deploying generative models for text generation, gene prediction, mathematical problem-solving, etc., producing outputs that adhere to intricate commonsense, biological, or mathematical logics in $\mathcal{KB}$.

\subsection{Brief Introduction to Abductive Learning}
\label{zoopt}

\begin{figure}
    \includegraphics[width=0.93\linewidth]{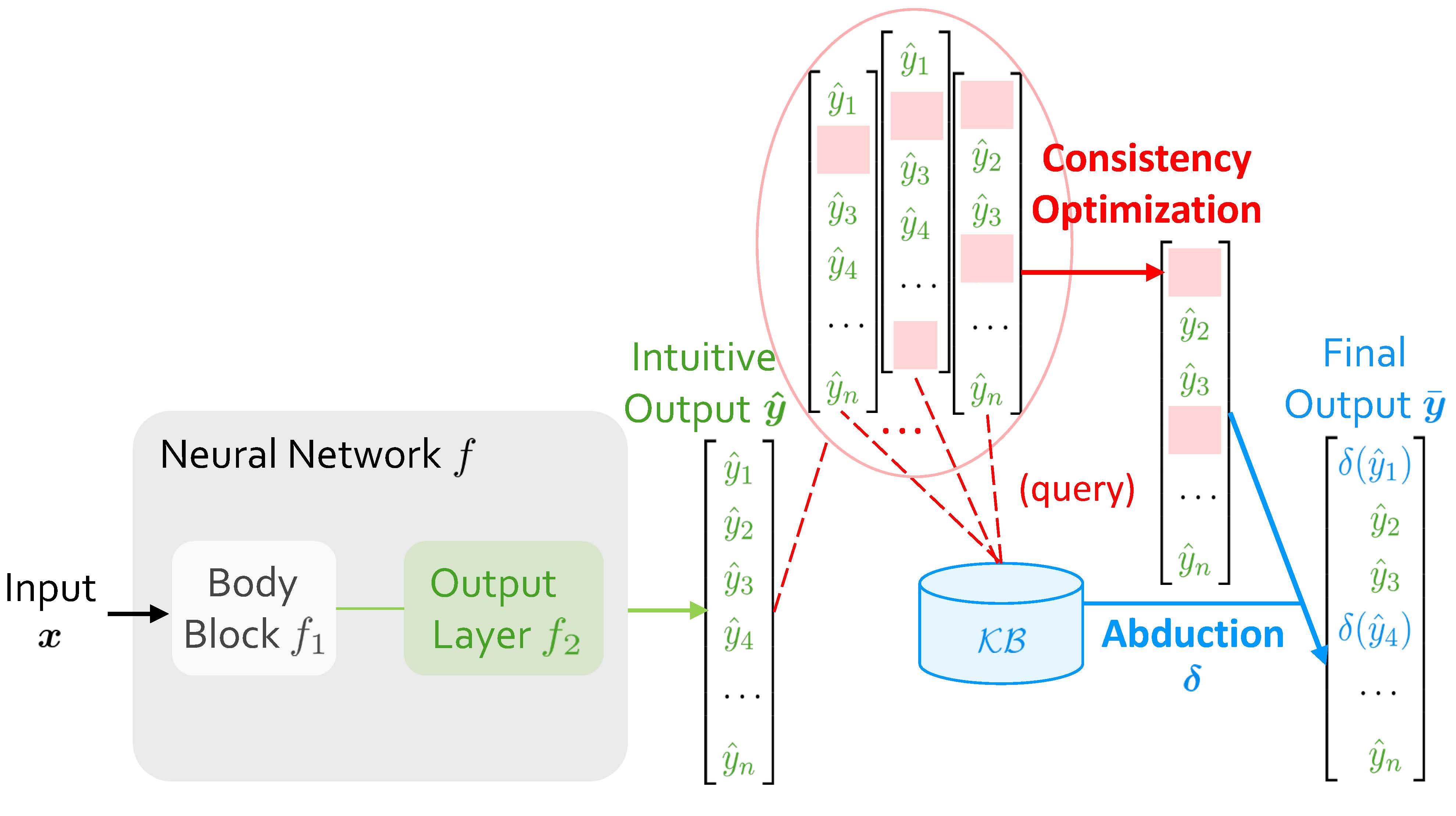}
    \caption{Abductive Learning (ABL) framework.}
    \label{ABL}
\end{figure}

When Abductive Learning (ABL) receives an input $\boldsymbol{x}$, it initially employs $f$ to map $\boldsymbol{x}$ into an intuitive output $\boldsymbol{\hat{y}} = \left[\hat{y}_1, \hat{y}_2,
\dots, \hat{y}_n\right]$. When $f$ is under-trained, $\boldsymbol{\hat{y}}$ might contain errors leading to inconsistencies with $\mathcal{KB}$. ABL then tries to rectify them, and obtains a revised $\boldsymbol{\bar{y}}$. As shown in Figure \ref{ABL}, the final output, $\boldsymbol{\bar{y}}$, consists of two parts: the green part retains the results from neural network, and the blue part is the modified result obtained by abduction, a basic form of symbolic reasoning that seeks plausible explanations for observations based on $\mathcal{KB}$.

Specifically, the process of obtaining $\boldsymbol{\bar{y}}$ can be divided into two sequential steps. The first step, consistency optimization, determines which positions in $\boldsymbol{\hat{y}}$ include elements that contain errors causing inconsistencies, so that performing abduction at these positions will yield a $\boldsymbol{\bar{y}}$ consistent with $\mathcal{KB}$. Essentially, this process is pinpointing propositions (or ground atoms, etc.) which have incorrect truth assignments, and most neuro-symbolic tasks can be formalized into this form. Once these positions are determined, the second step is rectifying by abduction, which then becomes easy for $\mathcal{KB}$ and its corresponding symbolic solver.

\paragraph{Challenge.}

In previous ABL, consistency optimization has always been a computational bottleneck. It operates as an external module using zeroth-order optimization methods, independent from both $f$ and $\mathcal{KB}$~\cite{dai2019bridging,zhou2022abl}. For each time of inference, it involves repetitively selecting various possible positions and querying the $\mathcal{KB}$ to see if a consistent result can be inferred. Each query involves an invocation of $\mathcal{KB}$ for slow symbolic reasoning. Also, since it is a complex combinatorial problem with a highly discrete nature, the number of such queries required escalates exponentially as data scale increases. This large number leads to a marked increase in time consumption, hence confines the applicability of ABL to only small datasets, usually those with output dimension $n$ less than 10.

\subsection{Architecture}

To address the challenges above, we propose Abductive Reflection (ABL-Refl). In this section, we will provide a detailed description of its architecture.

Let's first revisit the role of the neural network $f$ when we map the input to symbols from the set $\mathcal{Y}$. Typically, the raw data is first passed through the body block of the network, denoted by $f_1$, resulting in a high-dimensional embedding which encapsulates a wealth of feature information of the raw data. The form of $f_1$ varies, including structures like recurrent layers, graph convolution layers, or Transformers, etc. The result of $f_1$ is subsequently passed into several layers, usually linear layers, denoted by $f_2$, to obtain the intuitive output: $\boldsymbol{\hat{y}}=\text{argmax}(f_2(f_1(\boldsymbol{x})))\in\mathcal{Y}^n$.

\begin{figure}[t]
    \includegraphics[width=0.95\linewidth]{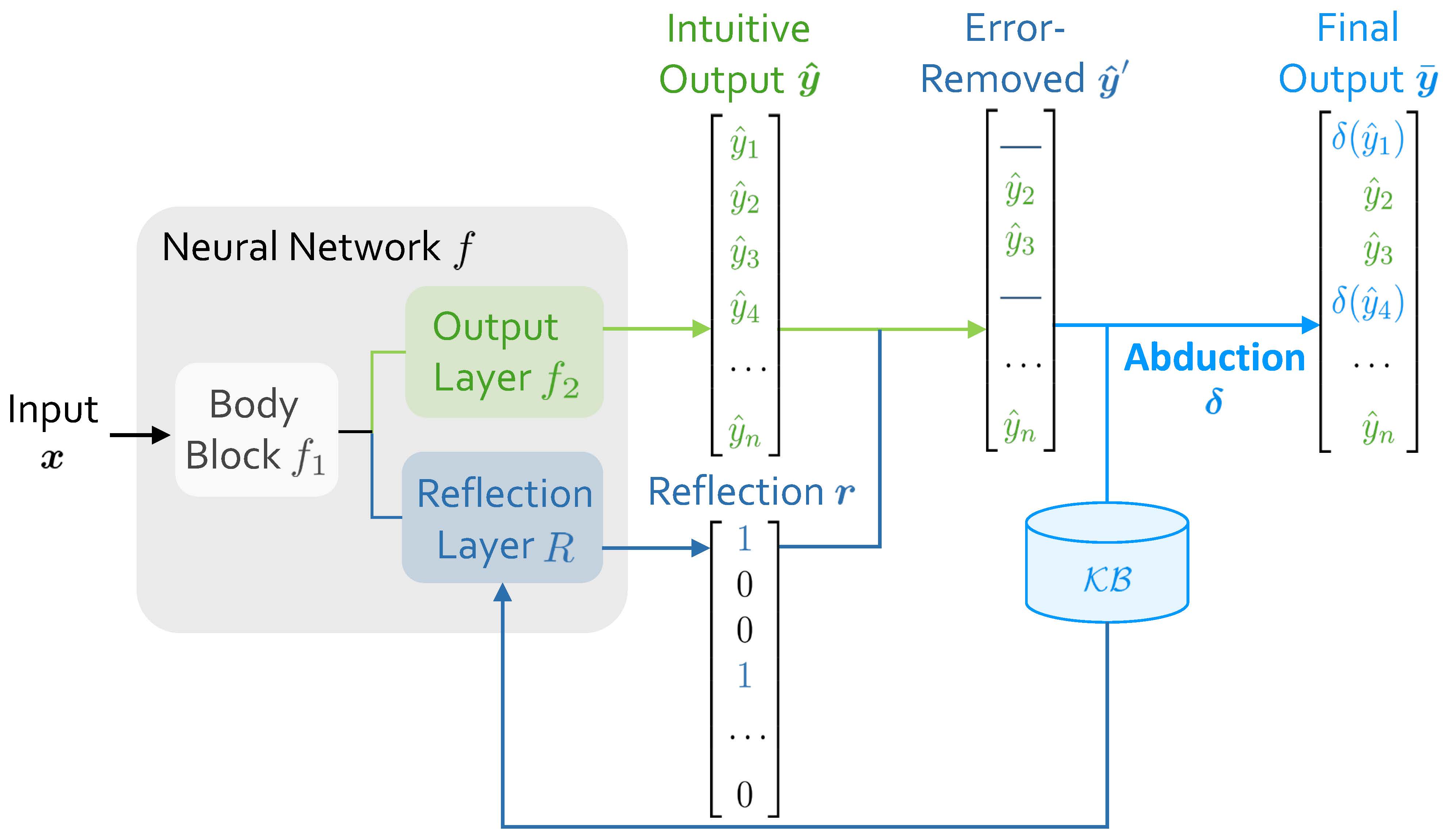}
    \caption{Architecture of Abductive Reflection (ABL-Refl). It replaces the external consistency optimization module with an efficient reflection mechanism, which is abduced directly from $\mathcal{KB}$.}
    \label{ABL-Refl}
\end{figure}

Besides the structure described above, as shown in Figure \ref{ABL-Refl}, our architecture further incorporates a reflection layer $R$ after the body block $f_1$, generating a reflection vector: $\boldsymbol{r}=\text{argmax}(R(f_1(\boldsymbol{x})))\in \{0,1\}^n$. The reflection layer $R$ and reflection vector $\boldsymbol{r}$ together constitute the reflection mechanism. This vector $\boldsymbol{r}$ has the same dimensionality $n$ as the intuitive output $\boldsymbol{\hat{y}}$, and each element, $r_i$, acts as a binary classifier to indicate whether the corresponding element $\hat{y}_i$ is an error leading to inconsistencies with $\mathcal{KB}$ (flagged as 1 for an error, and 0 otherwise). The reflection vector $\boldsymbol{r}$ is generated concurrently with the intuitive response during inference, resonating with human cognition where cognitive reflection typically forms right upon generation of an intuitive response~\cite{frederick2005cognitive}. 

With the initial intuitive output $\boldsymbol{\hat{y}}$ and the corresponding reflection vector $\boldsymbol{r}$, we seamlessly obtain the error-removed output $\hat{\boldsymbol{y}}^\prime$: In $\hat{\boldsymbol{y}}^\prime$, elements flagged as error by $\boldsymbol{r}$ are removed and left as blanks, while the rest are retained. Subsequently, $\mathcal{KB}$ applies abduction to fill in these blanks, thereby generating an output $\boldsymbol{\bar{y}}$ that is consistent with $\mathcal{KB}$. That is:
\begin{equation*}
    \bar{y}_i=\begin{cases}\quad\!\!\;\hat{y}_i, &r_i=0\\ \delta(\hat{y}_i), &r_i=1\end{cases}\quad i=1,2,\dots, n
    \label{y_hat'}
\end{equation*}
where $\delta$ denotes abduction. We treat $\boldsymbol{\bar{y}}= \left[\bar{y}_1, \bar{y}_2, \dots, \bar{y}_n\right]$ as the final output. 

During model training, the reflection is abduced from $\mathcal{KB}$ by directly leveraging information from domain knowledge (discussed later in Section \ref{training}). It can be seen as an attention mechanism generated from neural networks, which can help quickly focus symbolic reasoning specifically on areas it identifies as errors, hence largely narrowing the problem space of deliberate symbolic reasoning~\cite{zhang2020nlocalsat}. 

\paragraph{Benefits.}
Compared to previous ABL implementations, ABL-Refl replaces the zeroth-order consistency optimization module with the reflection mechanism to address the computational bottleneck. In this way, the need for a substantial number of querying $\mathcal{KB}$ is mitigated: After promptly pinpointing inconsistencies in System 1 output, regardless of the data scale, only a single invocation of $\mathcal{KB}$ is required to obtain a rectified and more consistent output.

Another thing worth noticing is that, in the architecture, the reflection layer directly connects to the body block, which helps leveraging information from the embeddings and linking more closely with the raw data. Therefore, the reflection vector $\boldsymbol{r}$ establishes a more direct and tighter bridge between raw data and domain knowledge.

\subsection{Training Paradigm}
\label{training}

In this section, we will discuss how to train the ABL-Refl method, especially the reflection in it.

\begin{figure}[t]
    \centering
    \includegraphics[width=0.65\linewidth]{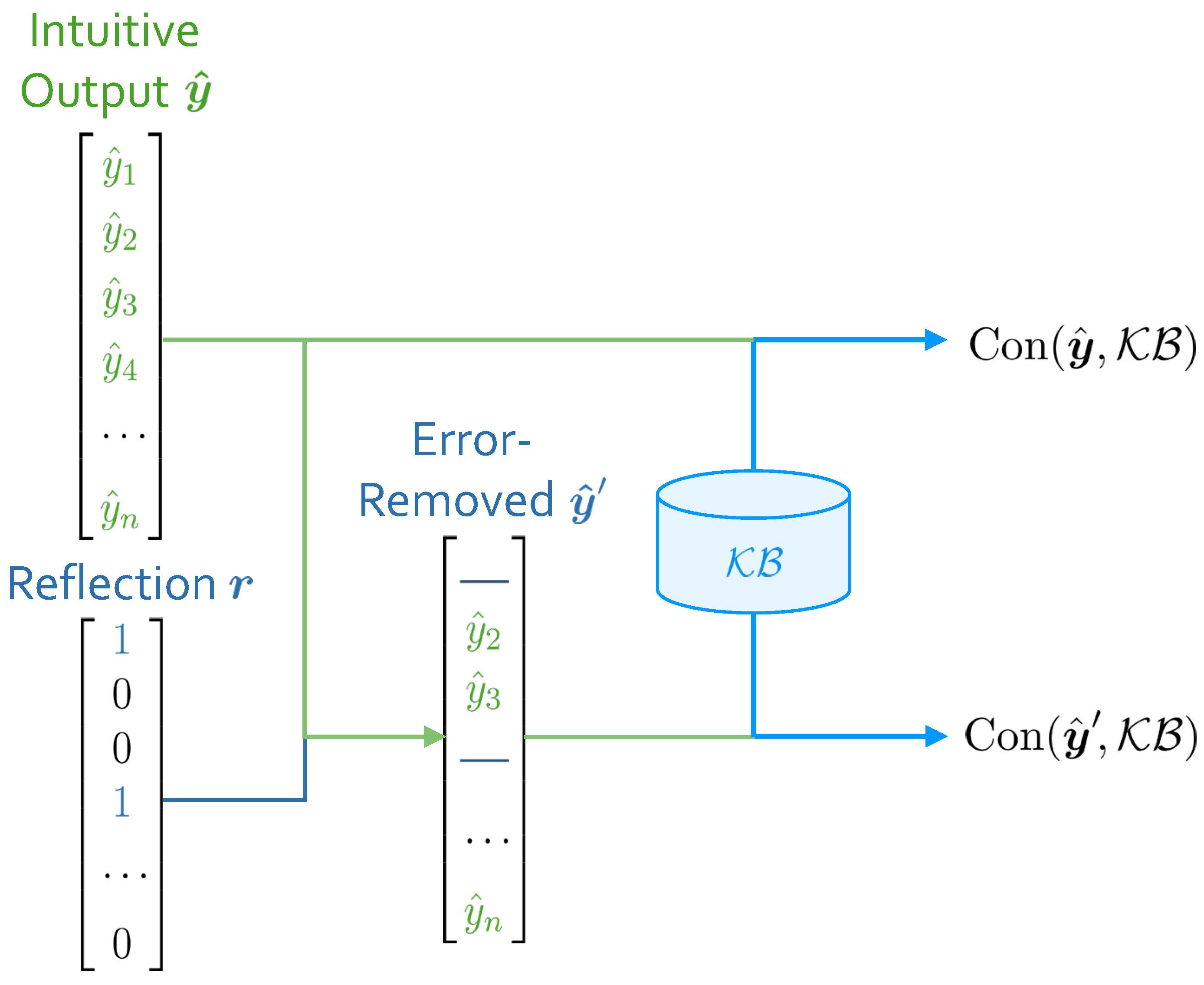}
    \caption{Consistency measurements.}
    \label{Opt}
\end{figure}

In ABL-Refl, when each input $\boldsymbol{x}$ is processed by the neural network, we obtain the intuitive output $\boldsymbol{\hat{y}}$ and the reflection vector $\boldsymbol{r}$, and subsequently obtain the error-removed (by $\boldsymbol{r}$) output $\boldsymbol{\hat{y}}^\prime$. With $\boldsymbol{\hat{y}}$ and $\boldsymbol{\hat{y}}^\prime$, we can measure their consistency with $\mathcal{KB}$, respectively. We denote these consistency measurements as $\text{Con}(\boldsymbol{\hat{y}}, \mathcal{KB})$ and $\text{Con}(\boldsymbol{\hat{y}}^\prime, \mathcal{KB})$, as shown in Figure \ref{Opt}. For a simplest example, if all elements in $\boldsymbol{\hat{y}}$ (or $\boldsymbol{\hat{y}}^\prime$) adhere to constraints in $\mathcal{KB}$, the consistency measurement is 1; otherwise, it is 0. 

Consequently, the improvement in consistency measurement after reflection, as denoted by
\begin{equation*}
    \Delta\text{Con}_{\boldsymbol{r}}(\boldsymbol{\hat{y}})=\text{Con}(\boldsymbol{\hat{y}}', \mathcal{KB}) - \text{Con}(\boldsymbol{\hat{y}}, \mathcal{KB})
\end{equation*}
naturally indicates the effectiveness of the reflection vector: A higher value of it signifies that reflection $\boldsymbol{r}$ can more effectively detect inconsistencies within $\boldsymbol{\hat{y}}$. Our training goal is to guide the neural network's parameters towards generating reflections that can maximize this value. Given that $\Delta\text{Con}_{\boldsymbol{r}}(\boldsymbol{\hat{y}})$ is usually a discrete value, we employ the REINFORCE algorithm to achieve this goal~\cite{williams1992simple}, which optimizes the policy (implicitly defined by neural network $f$) through maximizing a specified reward — in this case, $\Delta\text{Con}_{\boldsymbol{r}}(\boldsymbol{\hat{y}})$. This process leads to the following consistency loss:
\begin{equation}
    L_{con}(\boldsymbol{x})=-\Delta\text{Con}_{\boldsymbol{r}}(\boldsymbol{\hat{y}})\cdot\nabla_\theta\log f_\theta\left(\boldsymbol{\hat{y}}, \boldsymbol{r}\mid\boldsymbol{x}\right)
    \label{L_con}
\end{equation}
where $\theta$ are parameters of neural network $f$. 

Additionally, given that the time abduction required often escalates with problem size, we want to invoke it judiciously during inference, applying it only when it is truly necessary. Therefore, we aim to avoid the reflection vector from flagging too many elements in $\boldsymbol{\hat{y}}$ as error. To achieve this, we then introduce a reflection size loss:
\begin{equation}
    L_{size}(\boldsymbol{x})=\Phi\!\left(C-\frac{1}{n}\sum_{i=1}^n \left(1-R\left(f_1(\boldsymbol{x})\right)_i\right)\right)
    \label{L_size}
\end{equation}
where $\Phi(a)\triangleq \max(0,a)^2$ and $C$ is a hyperparameter ranging between 0 and 1. When $C$ is set at a higher value, the reflection vector tends to retain a greater number of intuitive output elements instead of flagging them as error and delegating to abduction.

In addition to the above-mentioned training methods, using labeled data, we employ data-driven supervised training methods similar to common neural network training paradigm. The loss function in this process, e.g., cross-entropy loss, is denoted by $L_{labeled}(\boldsymbol{x}, \boldsymbol{y})$. 

Therefore, combining all the training loss, the total loss for ABL-Refl is represented as follows:
\begin{equation}
    \begin{aligned}
        \mathcal{L}&=\frac{1}{|D_l|}\sum_{(\boldsymbol{x}, \boldsymbol{y})\in D_l} L_{labeled}(\boldsymbol{x}, \boldsymbol{y})\\
        &+\frac{1}{|D_l\cup D_u|}\sum_{\boldsymbol{x}\in D_l\cup D_u}(\alpha L_{con}(\boldsymbol{x}) + \beta L_{size}(\boldsymbol{x}) )
    \end{aligned}
    \label{L}
\end{equation} 
where $\alpha$ and $\beta$ are hyperparameters, $D_l=\{(\boldsymbol{x}_1, \boldsymbol{y}_1), (\boldsymbol{x}_2, \boldsymbol{y}_2), \dots\}$ are the labeled datasets and $D_u=\{\boldsymbol{x}_1, \boldsymbol{x}_2, \dots\}$ are the unlabeled datasets. 

Note that neither $L_{con}$ nor $L_{size}$, which are loss functions specifically related to the reflection, incorporate information from the data label. Instead, we leverage training information directly from $\mathcal{KB}$ to train the reflection. Also, despite sharing the prior feature layers, the output layer $f_2$ and reflection layer $R$ utilize different training information, thereby decoupling the objectives of intuitive problem-solving and inconsistency reflection.

\section{Experiments}
\label{exp}

In this section, we will conduct several experiments. First, we will test our method on the NeSy benchmark task of solving Sudoku to comprehensively verify its effectiveness. Next, we will change the Sudoku input from symbols to images, which requires integrating and simultaneous reasoning with both sub-symbolic and symbolic elements, representing one of the most challenging tasks in this field. Finally, we will tackle NP-hard combinatorial optimization problems on graphs, using a knowledge base of only mathematical definitions, to demonstrate our method's versatility. Through these experiments, we aim to answer the following questions:

\begin{itemize}
    \item[\textbf{Q1}] Compared to existing neuro-symbolic learning methods, can ABL-Refl achieve better performance in tasks requiring complex reasoning?

    \item[\textbf{Q2}] Can ABL-Refl reduce the training resources required?
    
    \item[\textbf{Q3}] Can ABL-Refl narrow the problem space for symbolic reasoning to achieve acceleration?
    
    \item[\textbf{Q4}] Does ABL-Refl possess the capability for broad application, such as handling diverse data scenarios or various forms of domain knowledge?
\end{itemize}

All experiments are performed on a server with Intel Xeon Gold 6226R CPU and Tesla A100 GPU. In our experiments, we simply set hyperparameters $\alpha$ and $\beta$ in Eq. \eqref{L} to 1, since adjusting them does not have a noticeable impact on the results. For the hyperparameter $C$ in \eqref{L_size}, we set it to 0.8, and have provided discussions in Appendix \ref{app_c}, demonstrating that setting it to a value within a broad moderate range (e.g., 0.6-0.9) would always be a recommended choice. All experiments are repeated 5 times.

\subsection{Solving Sudoku}
\label{exp1}

\paragraph{Dataset and Setting.} This task aims to solve a 9$\times$9 Sudoku: Given 81 digits of 0-9 (where 0 represents a blank space) in a 9$\times$9 board, we aim to find a solution $\boldsymbol{y}\in\{1,2,\dots,9\}^{81}$ that adhere to the Sudoku rules: no duplicate numbers are allowed in any row, column, or 3$\times$3 subgrid. In this section, we first consider inputs in symbolic form, $\boldsymbol{x}\in\{0,1,\dots,9\}^{81}$, and use datasets from a publicly available Kaggle site~\cite{rao_sudoku}.

For the neural network $f$, we use a simple graph neural network (GNN): the body block $f_1$ consists of one embedding layer and eight iterations of message-passing layers, resulting in a 128-dimensional embedding for each number, and then connects to both a linear output layer $f_2$ to obtain the intuitive output $\hat{\boldsymbol{y}}$ and a linear reflection layer $R$ to obtain the reflection vector ${\boldsymbol{r}}$. We use the cross-entropy loss as $L_{labeled}$. For the domain knowledge base $\mathcal{KB}$, it contains the Sudoku rules mentioned above. We express $\mathcal{KB}$ in the form of propositional logic and utilize the MiniSAT solver~\cite{sorensson2010minisat}, an open-source SAT solver, as the symbolic solver to leverage $\mathcal{KB}$ and perform abduction.

For the consistency measurement, we define it as follows: one point is awarded for each row, each column and each 3$\times$3 subgrid with no duplicate numbers, additionally, ten points are awarded if the entire board has no inconsistencies with $\mathcal{KB}$. In this way, it is entirely based on $\mathcal{KB}$. Notice that we deviated from the 1 or 0 measurement example setup mentioned in Section \ref{training} to avoid a predominance of zero values in $\Delta\text{Con}_{\boldsymbol{r}}(\boldsymbol{\hat{y}})$ of Eq. \eqref{L_con}, facilitating effective training with the REINFORCE algorithm. Similar considerations are applied in subsequent experiments. 

\paragraph{Compared Methods and Results.} 

We compare ABL-Refl with the following baseline methods: 1) Recurrent Relational Network (RRN)~\cite{palm2018recurrent}, a pure neural network method, 2) CL-STE~\cite{yang2022injecting}, a representative method of logic-based regularized loss, and 3) SATNet~\cite{wang2019satnet}. A detailed description of these methods is provided in Appendix \ref{app_com}. We also report the result for Simple GNN, which is the very same neural network used in our setting, yet directly treats the intuitive output $\hat{\boldsymbol{y}}$ as the final output. 

\begin{table}[t]
    \centering
    \begin{tabular}{lccc}
    \toprule
    \multicolumn{1}{c}{\textbf{Method}}  & 
    \textbf{\begin{tabular}[c]{@{}c@{}}Training\\ Time (min)\end{tabular}}  & \textbf{\begin{tabular}[c]{@{}c@{}}Inference\\ Time (s)\end{tabular}}
    & \textbf{\begin{tabular}[c]{@{}c@{}}Inference\\ Accuracy\end{tabular}} \\
    \midrule
    RRN            &   114.8\scriptsize{$\pm$7.8}  & 0.19\scriptsize{$\pm$0.01} & 73.1\scriptsize{$\pm$1.2}           \\
    CL-STE     &   173.6\scriptsize{$\pm$9.9}   &  0.19\scriptsize{$\pm$0.02}  & 76.5\scriptsize{$\pm$1.8}          \\
    SATNet            &   140.3\scriptsize{$\pm$6.8}   &   0.11\scriptsize{$\pm$0.01} & 74.1\scriptsize{$\pm$0.4}              \\
    \textbf{ABL-Refl}         &  \textbf{109.8\scriptsize{$\pm$10.8}} &   0.22\scriptsize{$\pm$0.02}  &  \textbf{97.4\scriptsize{$\pm$0.3}}        \\
    \midrule
    Simple GNN     &   29.7\scriptsize{$\pm$2.6}  & 0.02\scriptsize{$\pm$0.00}  & 55.6\scriptsize{$\pm$0.3}           \\
    \bottomrule
    \end{tabular}
    \caption{Training time (for a total of 100 epochs using 20K training data), inference time and accuracy (on 1K test data) on solving Sudoku.}
    \label{result_sudoku}
\end{table}

We report the training time (for a total of 100 epochs using 20K training data), inference time (on 1K test data) and accuracy (the percentage of completely accurate Sudoku solution boards on test data) in Table \ref{result_sudoku}. We may see that our method outperforms the baselines significantly, improving by over 20\% while maintaining a comparable inference time. This suggests an answer to \textbf{Q1}: ABL-Refl can achieve better reasoning performance. This improvement is primarily due to the use of abduction to rectify the neural network's output during inference. 

Furthermore, our method reaches high accuracy in only a few epochs (training curve is shown in Appendix \ref{app_training_curve}), significantly reducing training time. Even considering under identical training epochs, our total training time is less than baseline methods, despite involving a time-consuming symbolic solver. This partly stems from the neural network in our approach being less complex than those in baseline methods while achieving high accuracy. Overall, this suggests an answer to \textbf{Q2}: ABL-Refl can reduce the training time required.

We also attempt to reduce the amount of labeled data, removing labels from 50\%, 75\%, and 90\% of the training data. We record the inference accuracy in Table \ref{result_data}. It can be observed that even with only 2K labeled training data, our method still achieves far better accuracy than the baseline methods with 20K labeled training data. This suggests an answer to \textbf{Q2} from another aspect: ABL-Refl can reduce the labeled training data required.

\begin{table}[t]
    \centering
    \begin{tabular}{ccc}
    \toprule
    \textbf{Labeled Data} & \textbf{Unlabeled Data}  &  \textbf{Inference Accuracy} \\ 
    \midrule
    20K          & 0              &    97.4\scriptsize{$\pm$0.3}          \\
    10K          & 10K            &    96.3\scriptsize{$\pm$0.3}             \\
    5K          & 15K           &    95.8\scriptsize{$\pm$0.6}             \\
    2K           & 18K           &    94.7\scriptsize{$\pm$0.8}            \\ \bottomrule
    \end{tabular}
    \caption{Inference accuracy on solving Sudoku after reducing the amount of labeled data.}
    \label{result_data}
\end{table}

\paragraph{Comparing to Symbolic Solvers.}

\begin{table*}
    \centering
    \begin{tabular}{cclcccc}
    \toprule
    \multicolumn{1}{c}{\multirow{2}{*}{$\boldsymbol{\mathcal{KB}}$ \textbf{Form}}} & \multicolumn{1}{c}{\multirow{2}{*}{\textbf{Solver}}} & \multicolumn{1}{c}{\multirow{2}{*}{\textbf{Method}}} & \multirow{2}{*}{\textbf{\begin{tabular}[c]{@{}c@{}}Inference\\ Accuracy\end{tabular}}} & \multicolumn{3}{c}{\textbf{Inference Time (s)}} \\ 
    \cmidrule(r){5-7}
     &  &  &  &  \textbf{NN Time} & \textbf{Abduction Time} & \textbf{Overall Time} \\
    \midrule
    \multirow{2}{*}{\begin{tabular}[c]{@{}c@{}}Propositional\\ logic\end{tabular}} & \multirow{2}{*}{MiniSAT} & Solver only & 100\scriptsize{$\pm$0} & - & 0.227\scriptsize{$\pm$0.024} & 0.227\scriptsize{$\pm$0.024}  \\ 
    \cmidrule(r){3-7} 
    & & \textbf{ABL-Refl} & 97.4\scriptsize{$\pm$0.3} & 0.021\scriptsize{$\pm$0.004} & \textbf{0.196\scriptsize{$\pm$0.015}} & \textbf{0.217\scriptsize{$\pm$0.019}}  \\
    \midrule
    \multirow{2}{*}{\begin{tabular}[c]{@{}c@{}}First-order\\ logic\end{tabular}} & \multirow{2}{*}{\begin{tabular}[c]{@{}c@{}}Prolog with\\ CLP(FD)\end{tabular}} & Solver only & 100\scriptsize{$\pm$0} & - & 105.81\scriptsize{$\pm$5.62} & 105.81\scriptsize{$\pm$5.62}  \\
    \cmidrule(r){3-7} 
    & & \textbf{ABL-Refl} & 97.4\scriptsize{$\pm$0.3} & 0.021\scriptsize{$\pm$0.004} & \textbf{31.86\scriptsize{$\pm$1.88}} & \textbf{31.88\scriptsize{$\pm$1.89}}  \\
    \bottomrule
    \end{tabular}
    \caption{Inference accuracy and time (on 1K test data) on solving Sudoku. For $\mathcal{KB}$ expressed in two different forms, ABL-Refl shows notable acceleration compared to symbolic solvers in both cases.}
    \label{result_C}
\end{table*}

We next compare our method with merely employing symbolic solvers from scratch, to demonstrate its capability in accelerating symbolic reasoning. We perform inference on 1K test data and record the accuracy and time in Table \ref{result_C}. The inference time for our method includes the combined duration for data processing through both the neural network (NN time) and symbolic reasoning (abduction time). 

As observed in the former two lines, our method achieves a notable acceleration in the abduction process, consequently decreasing the overall inference time, with only a minor compromise in accuracy. This efficiency gain is due to the fact that in ABL-Refl, after quickly generating an intuition through the neural network, abduction only needs to focus on areas identified as necessary by the reflection vector, whereas using only symbolic solvers requires abduction to reason through all blanks in a Sudoku puzzle. Overall, this suggests an answer to \textbf{Q3}: ABL-Refl can quickly generate the reflection, thereby reducing the symbolic reasoning search space and enhancing reasoning efficiency.

We also compared with Prolog with CLP(FD)~\cite{triska2012finite} solver, by expressing the same $\mathcal{KB}$ with a first-order constraint logic program. As shown in the table, we observe a significant reduction in abduction time and overall inference time, which puts another evidence to our previous answer to \textbf{Q3}, and also suggests an answer to \textbf{Q4}: ABL-Refl can effectively utilize the two most commonly used forms in symbolic knowledge representation, propositional logic and first-order logic.

\subsection{Solving Visual Sudoku}
\label{exp2}

\paragraph{Dataset and Setting.} In this section, we modify the input from 81 symbolic digits to 81 MNIST images (handwritten digits of 0-9). We use the dataset provided in SATNet~\cite{wang2019satnet} and use 9K Sudoku boards for training and 1K for testing. 

In order to process image data, we first pass each image through a LeNet convolutional neural network (CNN)~\cite{lecun1998gradient} to obtain the probability of each digit. The rest of our setting follows from that described in Section \ref{exp1}.

\paragraph{Compared Methods and Results.}
We compare ABL-Refl with SATNet, as both methods allow for end-to-end training from visual inputs. We report the results in Table \ref{result_visual_sudoku} and the training curve in Appendix \ref{app_training_curve}. Compared to SATNet, ABL-Refl shows notable improvement in reasoning accuracy within only a few training epochs. We then consider pretraining the CNN in advance using self-supervised learning methods~\cite{chen2020simple} and find that this can further improve accuracy. Overall, the results further suggest positive answers to \textbf{Q1} and \textbf{Q2}. 

We also compare with CNN+Solver: each image is first mapped to symbolic form by a fully trained CNN (with 99.6\% accuracy on the MNIST dataset) and then directly fed into the symbolic solver to fill in the blanks and derive the final output. In such scenarios, the problem space for the symbolic solver includes all the Sudoku blanks, and additionally, since the symbolic solver cannot revise errors from CNN, any inaccuracies in CNN's output could lead the symbolic solver to crash (i.e., output no solution). Consequently, inference accuracy and time are adversely affected. This confirms the positive answer to \textbf{Q3}. 

Finally, an overview of Sections \ref{exp1} and \ref{exp2} also suggests an answer to \textbf{Q4}: ABL-Refl is capable of handling both symbolic and sub-symbolic forms of input data.

\begin{table}[t]
    \centering
    \begin{tabular}{lccc}
    \toprule
    \multicolumn{1}{c}{\textbf{Method}}  & \textbf{\begin{tabular}[c]{@{}c@{}}Inference\\ Time (s)\end{tabular}}  & 
    \textbf{\begin{tabular}[c]{@{}c@{}}Inference\\ Accuracy\end{tabular}}  \\ 
    \midrule
    SATNet              & 0.12\scriptsize{$\pm$0.01} & 63.5\scriptsize{$\pm$2.2}    \\
    CNN+Solver    & 0.23\scriptsize{$\pm$0.02} & 67.8\scriptsize{$\pm$4.2}      \\
    \textbf{ABL-Refl}      &  0.22\scriptsize{$\pm$0.02} & \textbf{77.8\scriptsize{$\pm$5.8}}    \\
    \textbf{ABL-Refl} \small{(with pretrained CNN)}  &  0.22\scriptsize{$\pm$0.02}  &   \textbf{93.5\scriptsize{$\pm$3.2}}         \\
    \bottomrule
    \end{tabular}
    \caption{Inference time (on 1K test data) and accuracy on solving visual Sudoku.}
    \label{result_visual_sudoku}
\end{table}

\subsection{Solving Combinatorial Optimization Problems on Graphs}
\label{exp3}

In this section, we will further expand the application domain of our method. We apply ABL-Refl to solving combinatorial optimization problems on graphs. We conduct the experiment on finding the maximum clique in this section, and provide an additional experiment in Appendix \ref{app_add}. 

\paragraph{Dataset and Setting.} In this task, we are given a graph $G=(V,E)$ with $|V|=n$ nodes, and aim to output $\boldsymbol{y}\in\{0,1\}^n$, where each index corresponds to a node, and the set of indices assigned the value of 1 collectively constitute the maximum clique. Note that this problem is a challenging NP-hard problem with extensive applications in real-life scenarios, and is generally considered challenging for neural networks~\cite{zhang2023rethinking}. 

We use several datasets from the TUDatasets~\cite{morris2020tudataset}, with their basic information shown in Table \ref{result_clique}. We use 80\% of the data for training and 20\% for testing.

In our method, the body layer $f_1$ consists of a single GAT layer~\cite{velivckovic2017graph} and 16 gated graph convolution layers~\cite{li2015gated}, and the output layer $f_2$ and reflection layer $R$ are both linear layers. We use binary cross-entropy loss as $L_{labeled}$. The domain knowledge base $\mathcal{KB}$ expresses the mathematical definition of maximum clique, i.e., every pair of vertices in the output set should be connected by an edge. We use Gurobi solver, an efficient mixed-integer program solver, to perform abduction. We define the consistency measurement as follows: one point is awarded for each pair of vertices if they are not connected by an edge; additionally, the size of the output set multiplied by 10 is added if the output set is indeed a clique.

\begin{table*}
    \centering
    \begin{tabular}{lcccc}
    \toprule
     \multicolumn{1}{c}{\multirow{2}{*}{\textbf{Method}}} & \multicolumn{4}{c}{\textbf{Dataset \footnotesize{(Graph nums./Avg. nodes per graph/Avg. edges per graph)}}} \\
    \cmidrule{2-5}
    \multicolumn{1}{c}{} & ENZYMES\footnotesize{ (600/33/62)} & PROTEINS\footnotesize{ (1113/39/73)} & IMDB-Binary\footnotesize{ (1000/19/97)} & COLLAB\footnotesize{ (5000/74/2457)} \\
    \midrule
    Erdos & 0.883\scriptsize{$\pm$0.156} &  0.905\scriptsize{$\pm$0.133} & 0.936\scriptsize{$\pm$0.175} & 0.852\scriptsize{$\pm$0.212} \\
    Neural SFE & 0.933\scriptsize{$\pm$0.148} & 0.926\scriptsize{$\pm$0.165} & 0.961\scriptsize{$\pm$0.143} & 0.781\scriptsize{$\pm$0.316} \\
    \textbf{ABL-Refl} & \textbf{0.991\scriptsize{$\pm$0.017}} & \textbf{0.985\scriptsize{$\pm$0.020}} & \textbf{0.979\scriptsize{$\pm$0.029}} & \textbf{0.982\scriptsize{$\pm$0.015}} \\
    \bottomrule
    \end{tabular}
    \caption{Approximation ratios on finding maximum clique on different datasets.}
    \label{result_clique}
\end{table*}

\paragraph{Compared Methods and Results.}
We compare our methods with the following baselines: 1) Erdos~\cite{karalias2020erdos}, 2) Neural SFE~\cite{karalias2022neural}, both leading methods for solving graph combinatorial problems. Their detailed descriptions are provided in Appendix \ref{app_com}. 

We report the approximation ratios in Table \ref{result_clique}. The approximation ratio, indicating the result set size relative to the actual maximum set size, is better when closer to 1. We may observe that our method outperforms the baseline methods, achieving near-perfect results on all datasets. This confirms the positive answer to \textbf{Q1}. Also, as the scale of the data increases, our method maintains a high level of accuracy, showing a more pronounced improvement compared to baseline methods. This suggests an answer to \textbf{Q4}: ABL-Refl is capable of handling scalable data scenarios, even in high-dimensional settings that are challenging for previous methods. Finally, an overview of this section provides another aspect to \textbf{Q4}: ABL-Refl can utilize a wide range of $\mathcal{KB}$, not limited to logical expressions but can also operate effectively with just the basic mathematical formulations.

\section{Effects of Reflection Mechanism}
\label{effects}

This section provides a further analysis on the reflection mechanism. In ABL-Refl, the reflection is abduced from domain knowledge, and acts as an efficient attention mechanism to direct the focus for symbolic search. This reflection is the key in our method to accomplish the NeSy reasoning rectification pipeline, i.e., a pipeline that detects errors in neural networks and then invokes symbolic reasoning to rectify these positions. To corroborate the effectiveness of the reflection, we conduct direct comparison with other methods that achieve the same pipeline: 
\begin{enumerate}
    \item[1)] \textbf{ABL}, minimizing the inconsistency of intuitive output and knowledge base with an external zeroth-order consistency optimization module, as detailed in Section \ref{zoopt}; 
    \item[2)] \textbf{NN Confidence}, retaining intuitive output with the top 80\% confidence from the neural network result (other retain thresholds are explored in Appendix \ref{app_nn}) and passing the remaining into symbolic reasoning; 
    \item[3)] \textbf{NASR}~\cite{cornelio2023learning}, using a Transformer-based external selection module to detect error, and the module is trained on a large synthetic dataset in advance. 
\end{enumerate}

We compare them on the solving visual Sudoku task in Section \ref{exp2}. For a fair comparison, all methods employ the same neural network, $\mathcal{KB}$ and MiniSAT solver setup. We report the recall (the percentage of errors from neural networks that can be identified), inference time and accuracy (on 1K test data) in Table \ref{result_selection}. Note that ``recall" directly evaluates the effectiveness of the detection module itself. The following analysis examines the results:

\begin{itemize}
    \item The consistency optimization in ABL faces significant efficiency challenges due to the large data scale (output dimension $n=81$). In such scenarios, the potential rectifications can reach up to $2^{81}$, resulting in an overwhelmingly large search space for consistency optimization. Also, as an external module, its only way of interacting with $\mathcal{KB}$ is to treat it as a black box and repetitively submit queries for consistency evaluation. As a result, it may require more than $10^9$ queries to identify errors for each Sudoku example, resulting in several hours to complete inference on 1K test data.

    \item NN Confidence performs poorly in identifying outputs with errors. Since the pure data-driven neural network training does not explicitly incorporate $\mathcal{KB}$ information, a low confidence from it does not necessarily indicate an inconsistency with the domain knowledge. This subsequently results in the frequent crashing in symbolic solver, therefore hampering the overall inference time and accuracy. This result parallels human cognitive reflection abilities, which do not show much positive correlation with System 1 intuition~\cite{pennycook2016cognitive}. To further illustrate this point, we provide additional analysis, including a case study, in Appendix \ref{app_nn}. 
    
    \item Our method also outperforms NASR, and notably, without the need of a synthetic dataset. This could be due to the fact that NASR's error-selection module is trained independently from other components, and operates sequentially and separately during inference. Therefore, it can only rely on information from the output label, in contrast to our method, which can leverage information directly from the body block of neural network, establishing a deeper connection with the raw data. Additionally, in NASR, traversing the separate selection module takes additional time, whereas in ABL-Refl, the reflection is generated concurrently with the neural network output, avoiding efficiency loss.
\end{itemize}

\begin{table}[t]
    \centering
    \begin{tabular}{lccc}
    \toprule
    \multicolumn{1}{c}{\textbf{Method}}  & \textbf{Recall} & \textbf{\begin{tabular}[c]{@{}c@{}}Inference\\ Time (s)\end{tabular}}  & 
    \textbf{\begin{tabular}[c]{@{}c@{}}Inference\\ Accuracy\end{tabular}}  \\ 
    \midrule
    ABL &  \multicolumn{1}{l}{Timeout}  & \multicolumn{1}{l}{Timeout}  &  \multicolumn{1}{l}{\;Timeout} \\
    NN Confidence &  82.64\scriptsize{$\pm$2.78} & 0.24\scriptsize{$\pm$0.03}  & 64.3\scriptsize{$\pm$6.2} \\
    NASR &  95.86\scriptsize{$\pm$0.96} & 0.26\scriptsize{$\pm$0.02}  & 82.7\scriptsize{$\pm$4.4} \\
    \textbf{ABL-Refl} &  \textbf{99.04\scriptsize{$\pm$0.85}} & \textbf{0.22\scriptsize{$\pm$0.02}}  & \textbf{93.5\scriptsize{$\pm$3.2}} \\
    \bottomrule
    \end{tabular}
    \caption{Recall, inference time and accuracy. "Timeout" indicates that inference takes more than 1 hour.}
    \label{result_selection}
\end{table}
 
\section{Conclusion}

In this paper, we present \textit{Abductive Reflection (ABL-Refl)}. It leverages domain knowledge to abduce a reflection vector, which flags potential errors in neural network outputs and then invokes abduction, serving as an attention mechanism for symbolic reasoning to focus on a much smaller problem space. Experiments show that ABL-Refl significantly outperforms other NeSy methods, achieving excellent reasoning accuracy with fewer training resources, and has successfully enhanced reasoning efficiency.

ABL-Refl preserves the integrity of both machine learning and logical reasoning with superior inference speed and high versatility. Therefore, it has the potential for broad application. In the future, it can be applied to large language models~\cite{mialon2023gaia} to help identify errors within their outputs, and subsequently exploit symbolic reasoning to enhance their trustworthiness and reliability.

\section*{Acknowledgments}
This research was supported by the NSFC~(62176117, 62206124) and Jiangsu Science Foundation Leading-edge Technology Program~(BK20232003).

\bibliography{aaai25}

\appendix
\onecolumn

\section{Comparison Methods}
\label{app_com}

In this section, we will provide a brief supplementary introduction to the compared baseline methods used in experiments.

\subsection{Solving Sudoku}

In the solving Sudoku experiment (Section \ref{exp1} and \ref{exp2}), we have compared our method with the following baselines: 

\begin{enumerate}
    \item[1)] \textbf{Recurrent Relational Network (RRN)}~\cite{palm2018recurrent}, a state-of-the-art pure neural network method tailored for this problem;
    \item[2)] \textbf{CL-STE}~\cite{yang2022injecting}, injecting logical knowledge (defined in the same way as our $\mathcal{KB}$) as neural network constraints during the training of RRN;
    \item[3)] \textbf{SATNet}~\cite{wang2019satnet}, incorporating a differentiable MaxSAT solver into the neural network to perform reasoning.
\end{enumerate}

Note that CL-STE is a representative method of logic-based regularized loss, relaxing symbolic logic as neural network loss. Additionally, among these methods, CL-STE stands out in both accuracy and efficiency (partly because it prevents constructing complex SDDs, unlike other methods including semantic loss~\cite{xu2018semantic}).

Other lines of methods generally underperform above baselines in scenarios where $n$ (the scale of $\boldsymbol{y}$) is high. For instance, \textbf{ABL} faces the challenge where consistency optimization needs to choose among exponential query candidates, resulting in runtimes thousands of times longer than other methods, as seen in Section \ref{effects}. Take two other representative NeSy methods as examples: \textbf{DeepProbLog}~\cite{manhaeve2018deepproblog} involves substantial computational costs, taking days to complete solving Sudoku; \textbf{NeurASP}~\cite{yang2020neurasp} also performs slow and lags in accuracy, as shown in Yang et al.~\shortcite{yang2022injecting}.

\subsection{Solving Combinatorial Optimization on Graphs}

In the solving combinatorial optimization on graphs experiment (Section \ref{exp3} and Appendix \ref{app_add}), we have compared our method with the following baselines:

\begin{enumerate}
    \item[1)] \textbf{Erdos}~\cite{karalias2020erdos}, optimizing set functions using a neural network parametrizing a distribution over sets;
    \item[2)] \textbf{Neural SFE}~\cite{karalias2022neural}, optimizing set functions by extending them onto high-dimensional continuous domains.
\end{enumerate}
In this experiment, the above methods use the same body block graph neural network as our method.

\section{Training Curve}
\label{app_training_curve}

\begin{figure}[h]
  \centering
  \begin{subfigure}[b]{0.495\textwidth}
      \includegraphics[width=\textwidth]{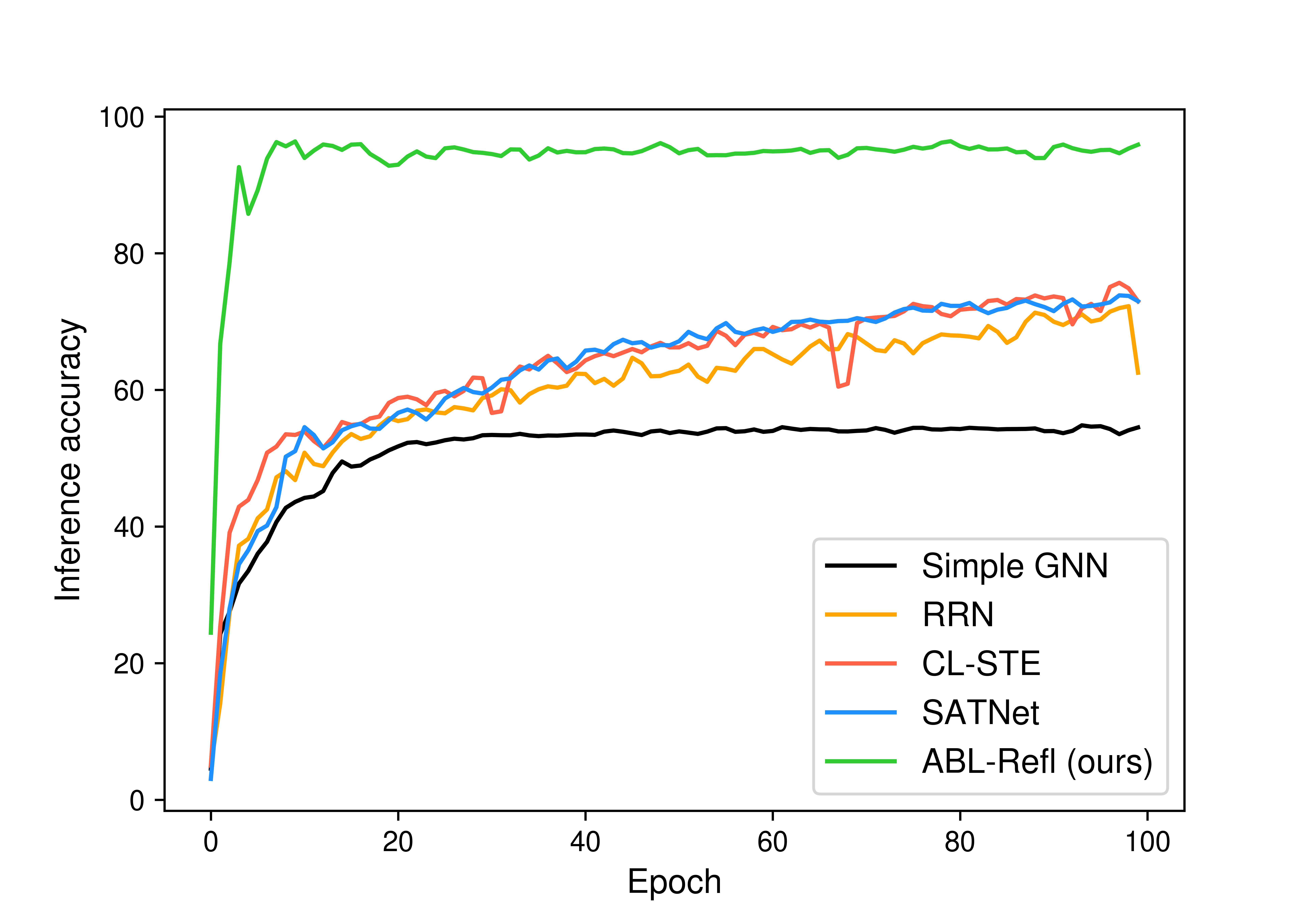}
      \caption{Sudoku.}
  \label{fig_sudoku}
  \end{subfigure}
  \hfill
  \begin{subfigure}[b]{0.495\textwidth}
    \includegraphics[width=\linewidth]{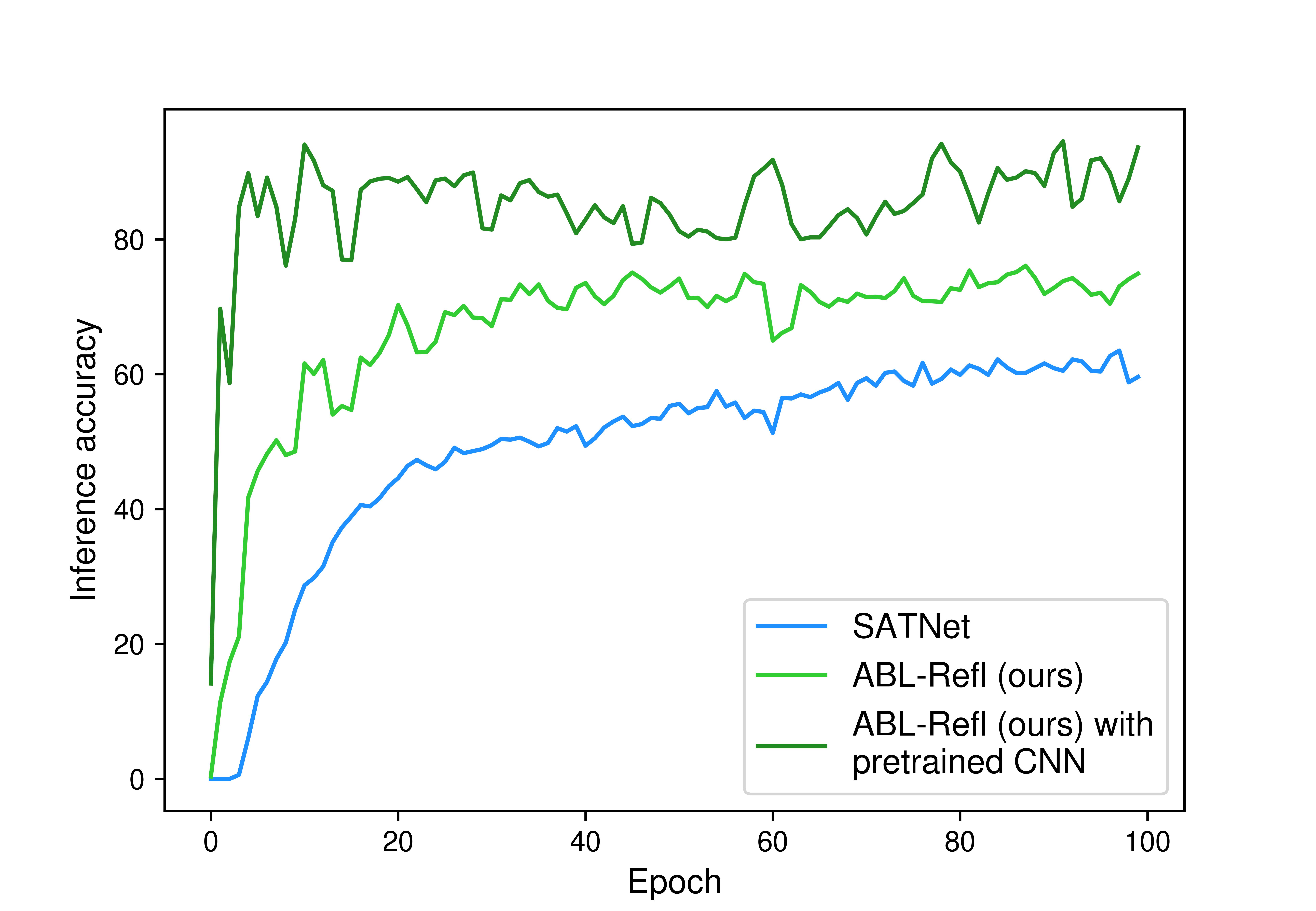}
    \caption{Visual Sudoku.}
    \label{fig_visual_sudoku}
  \end{subfigure}
  \caption{Training curve on solving Sudoku and visual Sudoku.}
\end{figure}

In this section, we will report the training curve in the experiments on solving Sudoku (Section \ref{exp1}) and visual Sudoku (Section \ref{exp2}). The respective training curves for each scenario are shown in Figures \ref{fig_sudoku} and \ref{fig_visual_sudoku}, with the horizontal axis representing training epochs and the vertical axis representing inference accuracy. We may see that our method achieves high accuracy within just a few epochs, significantly reducing training time compared to other baseline methods.

\section{Discussion on Hyperparameter $C$}
\label{app_c}

In this section, we will discuss the effect of the hyperparameter $C$. Previous experiments in Sections \ref{exp} and \ref{effects}, $C$ was consistently set to 0.8, and we will now explore adjustments. We report the extended results in Tables \ref{C_sudoku} and \ref{C_clique}. It is shown that when $C$ is set within a wide range, ABL-Refl uniformly outperforms the baseline methods. 

Intuitively, as mentioned in Section \ref{Opt}, setting $C$ lower delegates more elements to the solver for correction, thereby often enhancing reasoning accuracy. The results in Tables \ref{C_sudoku} and \ref{C_clique} have also demonstrated this point. 

\begin{table}[t]
    \centering
    \begin{tabular}{lcccc}
    \toprule
    \multicolumn{1}{c}{\textbf{Experiment}} & \multicolumn{2}{c}{\textbf{Method}} & \multicolumn{1}{c}{\textbf{Inference Time (s)}} & \multicolumn{1}{c}{\textbf{Inference Accuracy}} \\ 
    \midrule
    \multirow{7}{*}{Sudoku} & \multicolumn{2}{l}{Simple GNN} & 0.02\scriptsize{$\pm$0.00}  & 55.6\scriptsize{$\pm$0.3}   \\
     & \multicolumn{2}{l}{RNN} & 0.19\scriptsize{$\pm$0.01} & 73.1\scriptsize{$\pm$1.2} \\
     & \multicolumn{2}{l}{CL-STE} & 0.19\scriptsize{$\pm$0.02}  & 76.5\scriptsize{$\pm$1.8} \\
     & \multicolumn{2}{l}{SATNet} & 0.11\scriptsize{$\pm$0.01} & 74.1\scriptsize{$\pm$0.4}  \\ 
     \cmidrule{2-5} 
     & \multirow{3}{*}{\textbf{ABL-Refl}} & $C=0.7$ & 0.24\scriptsize{$\pm$0.02} & \textbf{99.1\scriptsize{$\pm$0.2}} \\
     &  & $C=0.8$ & 0.22\scriptsize{$\pm$0.02}  &  \textbf{97.4\scriptsize{$\pm$0.3}} \\
     &  & $C=0.9$ & 0.21\scriptsize{$\pm$0.02} & \textbf{96.6\scriptsize{$\pm$0.5}} \\ 
     \midrule
    \multicolumn{1}{l}{} & \multicolumn{2}{l}{SATNet}  & 0.12\scriptsize{$\pm$0.01} & 63.5\scriptsize{$\pm$2.2}  \\
    \multicolumn{1}{l}{\multirow{5}{*}{Visual Sudoku}} & \multicolumn{2}{l}{CNN+Solver}  & 0.23\scriptsize{$\pm$0.02} & 67.8\scriptsize{$\pm$4.2}  \\
    \cmidrule{2-5} 
    \multicolumn{1}{l}{} & \multirow{3}{*}{\textbf{ABL-Refl}} & $C=0.7$ & 0.24\scriptsize{$\pm$0.02} & \textbf{95.9\scriptsize{$\pm$2.8}} \\
    \multicolumn{1}{l}{} &  & $C=0.8$ & 0.22\scriptsize{$\pm$0.02} & \textbf{93.5\scriptsize{$\pm$3.2}}  \\
    \multicolumn{1}{l}{} &  & $C=0.9$ & 0.21\scriptsize{$\pm$0.02} & \textbf{90.6\scriptsize{$\pm$4.2}} \\
    \bottomrule
    \end{tabular}
    \caption{Inference time and accuracy on solving Sudoku and visual Sudoku. For different values of the hyperparameter $C$, ABL-Refl uniformly outperforms other baseline methods.}
    \label{C_sudoku}
\end{table}

\begin{table}[t]
    \centering
    \begin{tabular}{lccccc}
    \toprule
    \multicolumn{2}{c}{\multirow{2}{*}{\textbf{Method}}} & \multicolumn{4}{c}{\textbf{Dataset }} \\
    \cmidrule{3-6}
    \multicolumn{2}{c}{} & ENZYMES & PROTEINS & IMDB-Binary & COLLAB \\
    \midrule
    \multicolumn{2}{l}{Erdos} & 0.883\scriptsize{$\pm$0.156} &  0.905\scriptsize{$\pm$0.133} & 0.936\scriptsize{$\pm$0.175} & 0.852\scriptsize{$\pm$0.212} \\
    \multicolumn{2}{l}{Neural SFE} & 0.933\scriptsize{$\pm$0.148} & 0.926\scriptsize{$\pm$0.165} & 0.961\scriptsize{$\pm$0.143} & 0.781\scriptsize{$\pm$0.316} \\
    \midrule
    \multirow{3}{*}{\textbf{ABL-Refl}} & $C=0.7$ & \textbf{0.992\scriptsize{$\pm$0.012}} & \textbf{0.988\scriptsize{$\pm$0.019}} & \textbf{0.984\scriptsize{$\pm$0.026}} & \textbf{0.986\scriptsize{$\pm$0.016}} \\
    & $C=0.8$ & \textbf{0.991\scriptsize{$\pm$0.017}} & \textbf{0.985\scriptsize{$\pm$0.020}} & \textbf{0.979\scriptsize{$\pm$0.029}} & \textbf{0.982\scriptsize{$\pm$0.015}} \\
    & $C=0.9$ & \textbf{0.982\scriptsize{$\pm$0.023}} & \textbf{0.975\scriptsize{$\pm$0.021}} & \textbf{0.968\scriptsize{$\pm$0.035}} & \textbf{0.971\scriptsize{$\pm$0.021}} \\
    \bottomrule
    \end{tabular}
    \caption{Approximation ratios on finding maximum clique. For different values of the hyperparameter $C$, ABL-Refl uniformly outperforms other baseline methods.}
    \label{C_clique}
\end{table}

However, setting $C$ to more extreme lower values, while potentially further enhancing reasoning accuracy, will face the risk of weakening the reflection in accelerating reasoning, since more elements are delegated to symbolic reasoning. Therefore, we do not recommend excessively lowering $C$. For this effect of $C$ in computational efficiency, we have also conducted experimental evaluation: The runtime after adjusting $C$ are reported in Table \ref{C_solver}. It can be seen that setting $C$ to a higher value can further narrow the search space for symbolic reasoning, thereby offering a more substantial efficiency improvement. (On the contrary, setting $C$ to a more extreme high value would essentially rely merely on the neural network's intuitive output, rendering the reflection vector ineffective; hence, such settings are not considered.)

\begin{table}[t]
    \centering
    \begin{tabular}{cclccccc}
    \toprule
    \multicolumn{1}{c}{\multirow{2}{*}{$\boldsymbol{\mathcal{KB}}$ \textbf{Form}}} & \multicolumn{1}{c}{\multirow{2}{*}{\textbf{Solver}}} & \multicolumn{2}{c}{\multirow{2}{*}{\textbf{Method}}} & \multirow{2}{*}{\textbf{\begin{tabular}[c]{@{}c@{}}Inference\\ Accuracy\end{tabular}}} & \multicolumn{3}{c}{\textbf{Inference Time (s)}} \\ 
    \cmidrule(r){6-8}
     &  &  &  &  & \textbf{NN Time} & \textbf{Abduction Time} & \textbf{Overall Time} \\
    \midrule
    \multirow{4}{*}{\begin{tabular}[c]{@{}c@{}}Propositional\\ logic\end{tabular}} & \multirow{4}{*}{MiniSAT} & \multicolumn{2}{l}{Solver only} & 100\scriptsize{$\pm$0} & - & 0.227\scriptsize{$\pm$0.024} & 0.227\scriptsize{$\pm$0.024}  \\ 
    \cmidrule(r){3-8} 
    & & \multirow{3}{*}{\textbf{ABL-Refl}} & $C=0.7$ & 99.1\scriptsize{$\pm$0.2} & & 0.218\scriptsize{$\pm$0.019} & 0.239\scriptsize{$\pm$0.023} \\
    & & & $C=0.8$ & 97.4\scriptsize{$\pm$0.3} & 0.021\scriptsize{$\pm$0.004} & \textbf{0.196\scriptsize{$\pm$0.015}} & \textbf{0.217\scriptsize{$\pm$0.019}} \\
    & & & $C=0.9$ & 96.6\scriptsize{$\pm$0.5} &  & \textbf{0.185\scriptsize{$\pm$0.017}} &  \textbf{0.206\scriptsize{$\pm$0.021}} \\
    \midrule
    \multirow{4}{*}{\begin{tabular}[c]{@{}c@{}}First-order\\ logic\end{tabular}} & \multirow{4}{*}{\begin{tabular}[c]{@{}c@{}}Prolog with\\ CLP(FD)\end{tabular}} & \multicolumn{2}{l}{Solver only} & 100\scriptsize{$\pm$0} & - & 105.81\scriptsize{$\pm$5.62} & 105.81\scriptsize{$\pm$5.62}  \\
    \cmidrule(r){3-8} 
    & & \multirow{3}{*}{\textbf{ABL-Refl}} & $C=0.7$ & 99.1\scriptsize{$\pm$0.2} & & \textbf{68.59\scriptsize{$\pm$3.31}} & \textbf{68.61\scriptsize{$\pm$3.31}} \\
    & & & $C=0.8$ & 97.4\scriptsize{$\pm$0.3} & 0.021\scriptsize{$\pm$0.004} & \textbf{31.86\scriptsize{$\pm$1.88}} & \textbf{31.88\scriptsize{$\pm$1.89}} \\
    & & & $C=0.9$ & 96.6\scriptsize{$\pm$0.5} &  & \textbf{20.47\scriptsize{$\pm$1.23}} & \textbf{20.49\scriptsize{$\pm$1.23}} \\
    \bottomrule
    \end{tabular}
    \caption{Inference accuracy and time (on 1K test data) on solving Sudoku. Setting the hyperparameter $C$ to a higher value offers a more substantial efficiency improvement compared to symbolic solvers.}
    \label{C_solver}
\end{table}

In summary, to utilize the reflection vector’s role in bridging neural network outputs and symbolic reasoning, setting $C$ within a moderate range is advised. Experimental evidence suggests that within this broad range, e.g., 0.6-0.9, the specific value of $C$ actually does not significantly impact outcomes; it is merely a balance between accuracy and computation time.

\section{More Discussion on Comparison with Neural Network Confidence}
\label{app_nn}

The core idea of ABL-Refl is to identify areas in the neural network’s intuitive output where inconsistencies with knowledge are most likely to occur. Thus, a straightforward approach might seem to be letting the neural network itself highlight errors, i.e., treating elements with low confidence values from the neural network result as potential errors. However, Section \ref{effects} have proven that such a naive approach significantly underperforms our method. This is because neural networks cannot explicitly utilize symbolic knowledge during training, making it challenging to establish a correlation between confidence levels and inconsistencies with knowledge.

To illustrate this more clearly, we now demonstrate a case study in the solving Sudoku experiment: Figures \ref{case1}-\ref{case2} below depict a Sudoku problem and its correct solution. Figure \ref{case3} shows the intuitive output obtained from the GNN, where several numbers marked in red are incorrect. Figures \ref{case4} and \ref{case5} display the results using NN confidence and the reflection vector, respectively, with identified potential error positions in blue. 

\begin{figure}[h]
    \centering
    \begin{subfigure}[b]{0.25\textwidth}
        \includegraphics[width=\textwidth]{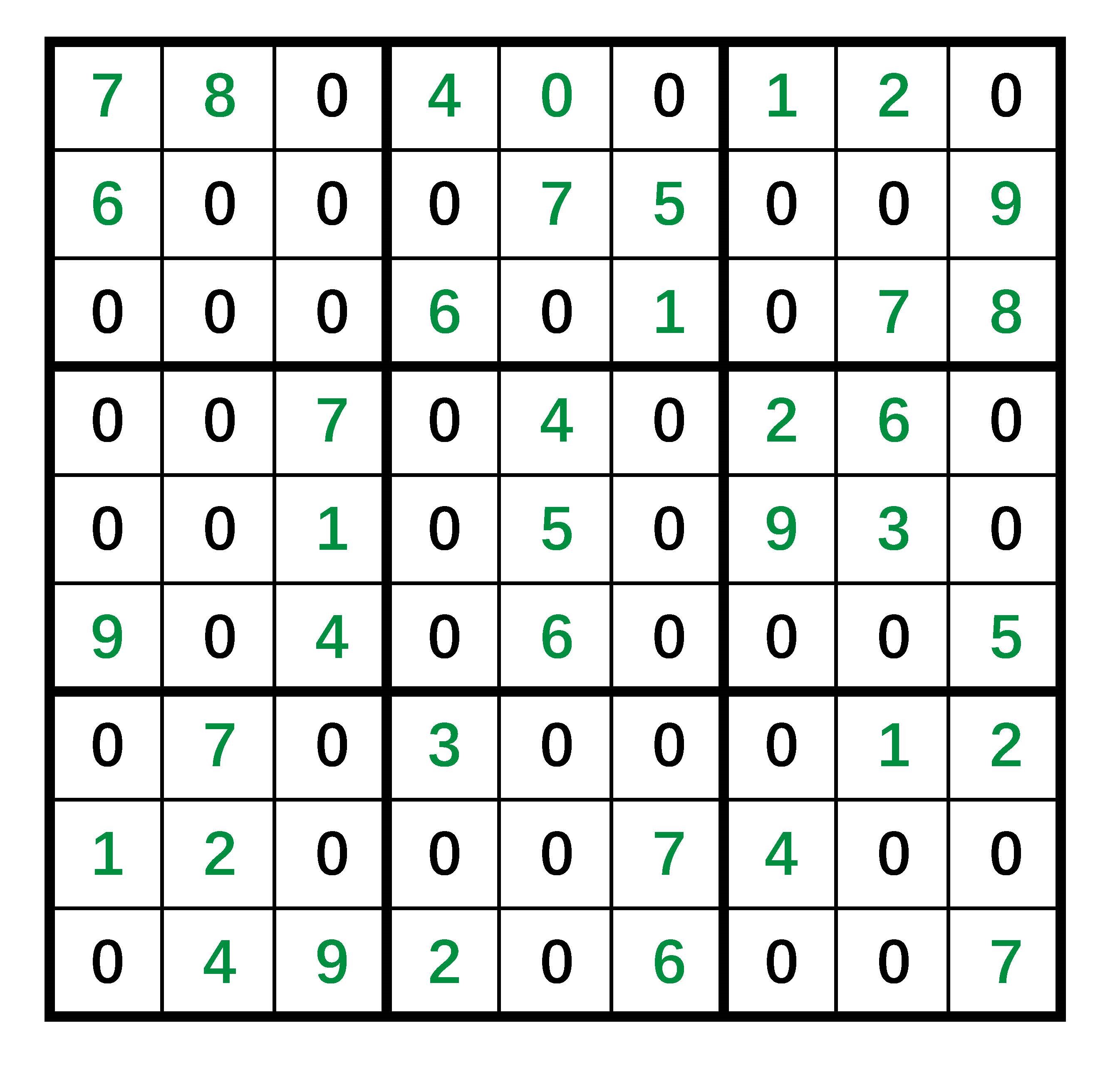}
        \caption{Sudoku problem}
        \label{case1}
    \end{subfigure}
    \hspace{25pt}
    \begin{subfigure}[b]{0.25\textwidth}
        \includegraphics[width=\textwidth]{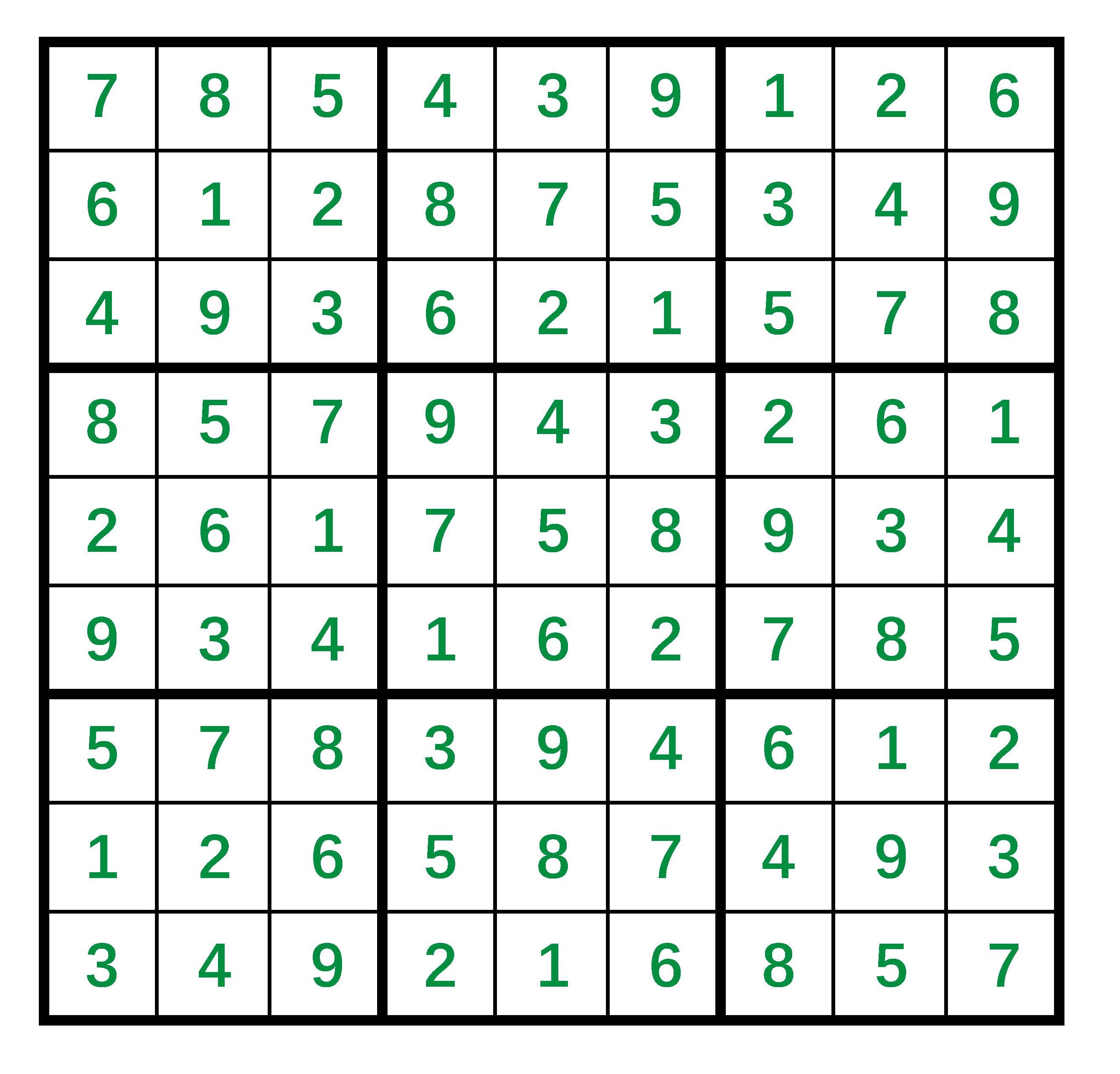}
        \caption{Sudoku solution}
        \label{case2}
    \end{subfigure}
    \hspace{25pt}
    \begin{subfigure}[b]{0.25\textwidth}
        \includegraphics[width=\textwidth]{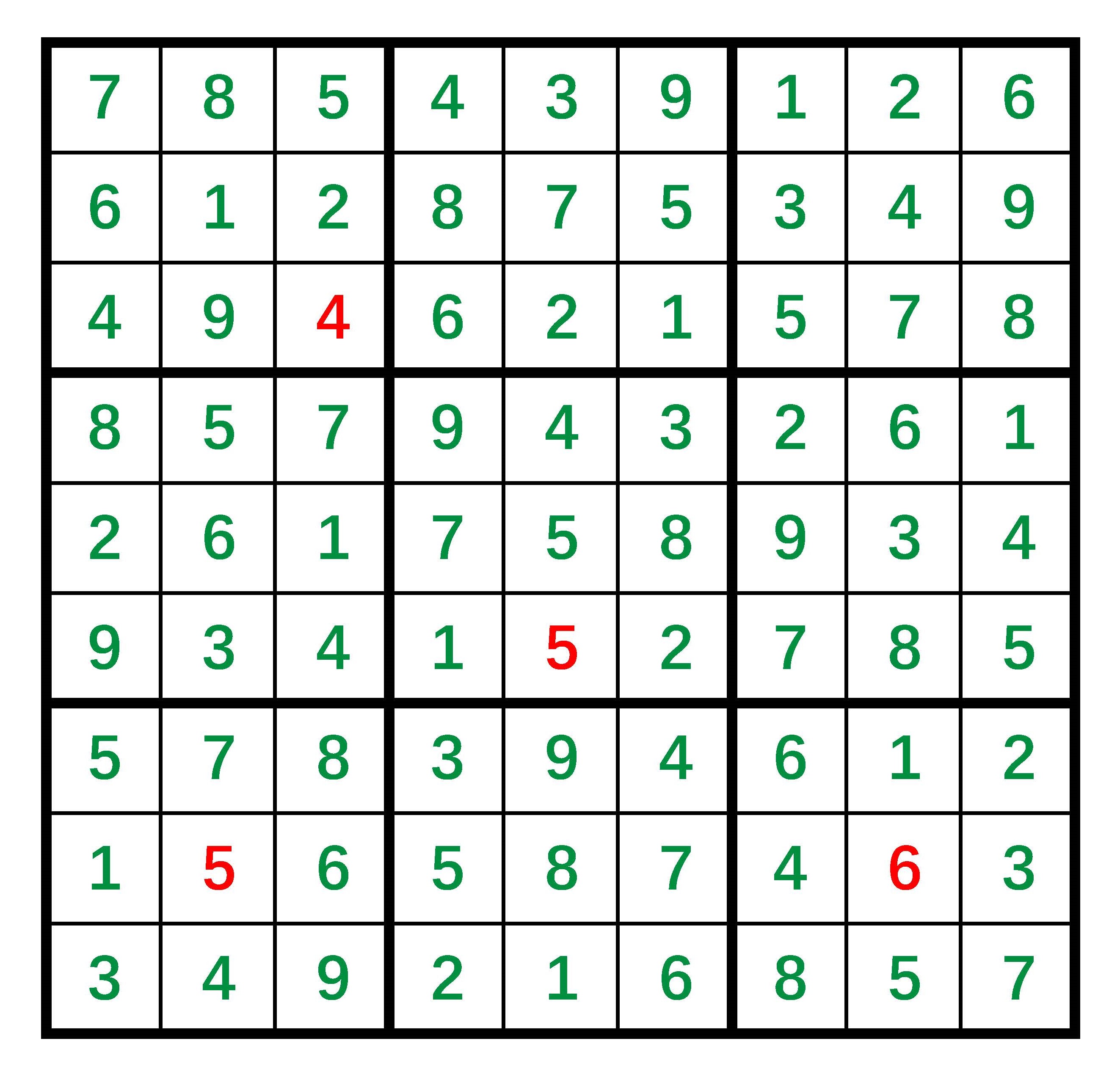}
        \caption{Neural network intuitive output}
        \label{case3}
    \end{subfigure}
    
    \begin{subfigure}[b]{0.333\textwidth}
        \centering
        \includegraphics[width=0.75\textwidth]{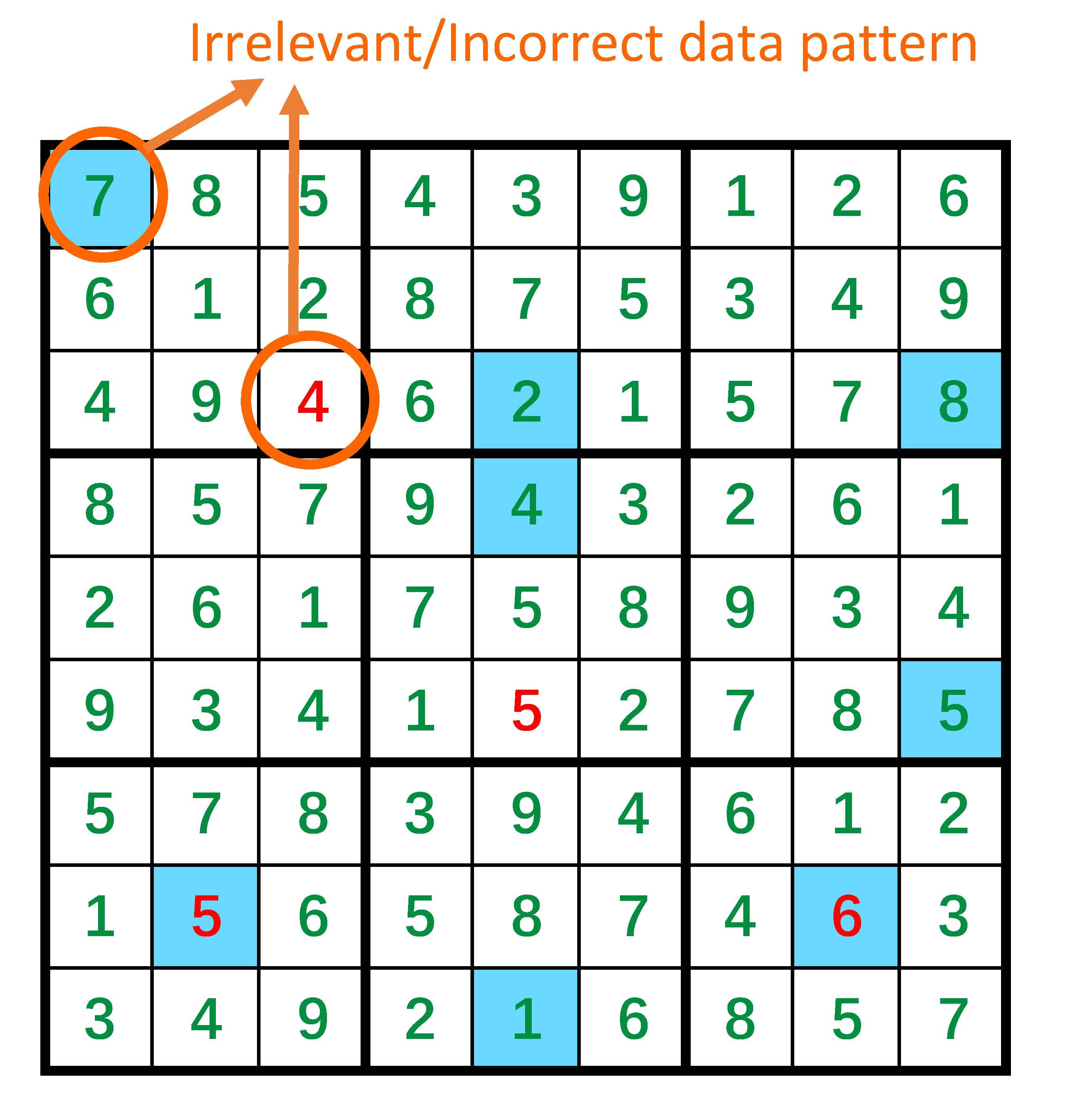}
        \caption{Errors identified by NN confidence}
        \label{case4}
    \end{subfigure}
    \hspace{5pt}
    \begin{subfigure}[b]{0.333\textwidth}
        \centering
        \includegraphics[width=0.75\textwidth]{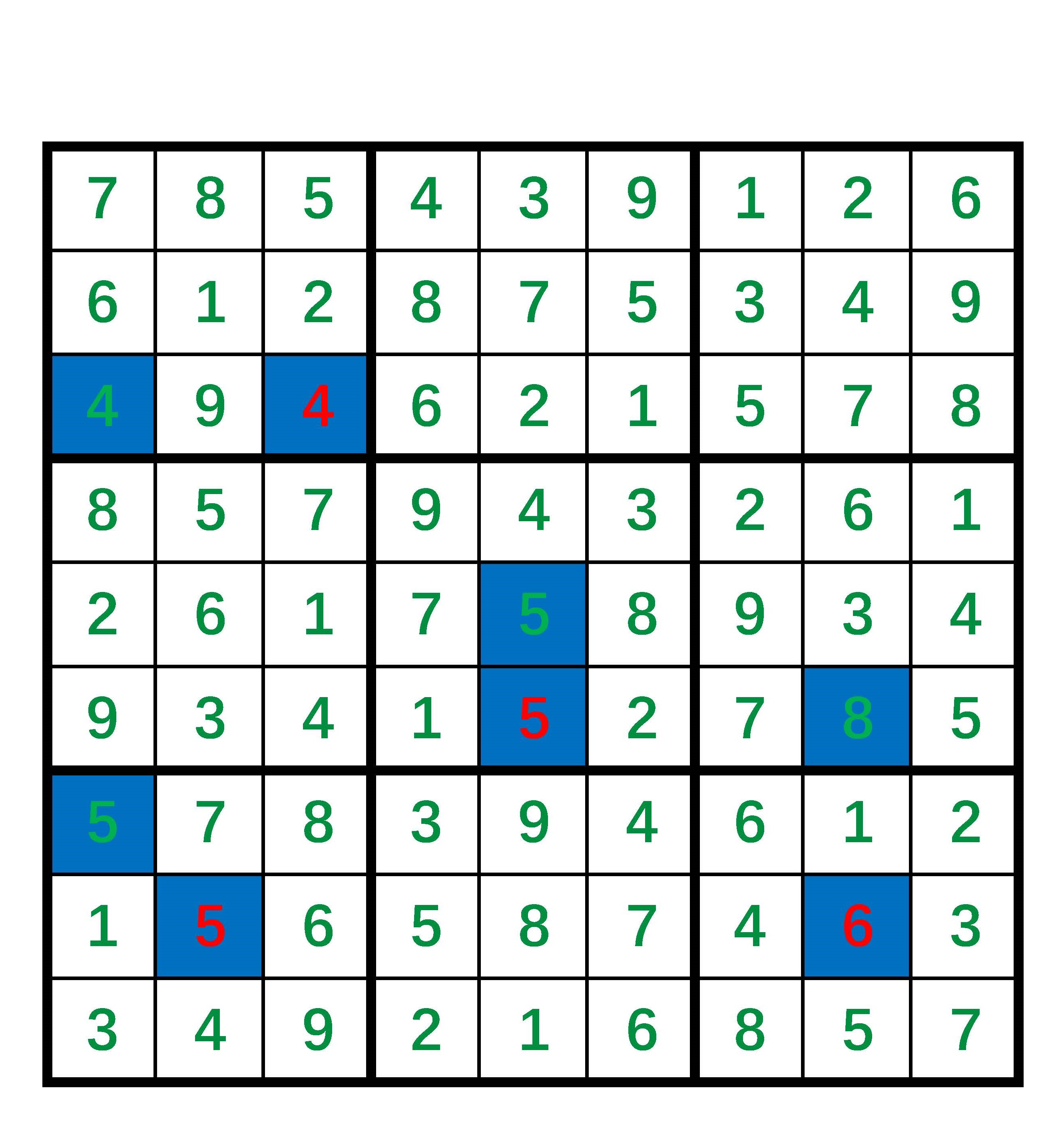}
        \caption{Errors identified by ABL-Refl}
        \label{case5}
    \end{subfigure}
    \caption{A case study in the solving Sudoku experiment.}
    \label{fig_case}
\end{figure}

\noindent It can be seen that the errors marked by the reflection vector generally correspond to the constraints in $\mathcal{KB}$, containing duplicate numbers either in a row, column, or subgrid. In contrast, errors identified by NN confidence are difficult to align with such knowledge. Take the incorrect identification of the first row, first column as an example, after examining the dataset, we find that there are some of the Sudoku solutions with a number ``4" in the third row, third column and a number ``7" in the first row, first column at the same time. These irrelevant yet common data patterns likely lead the neural network to erroneously learn during training. Hence, when an error occurs in the third row and third column, the confidence in the first row and first column also drops. This case study highlights that pure data-driven networks cannot explicitly utilize KB knowledge: during training, they only have access to data labels, not the logical principles behind the data. Consequently, due to factors like learning incorrect data patterns or overfitting to noise, confidence values often misalign with the compatibility with domain knowledge, leading them become unreliable to identify errors. In contrast, the training information for the reflection vector is directly derived from the $\mathcal{KB}$.

Furthermore, as discussed in Appendix \ref{app_c}, in ABL-Refl, adjusting the hyperparameter $C$, as a soft margin, can help determine how much of the neural network's output is retained. In Section \ref{effects}, corresponding to $C=0.8$, the neural network's output with top 80\% confidence was retained. We will now test adjusting this threshold of retaining neural network's output. We report the results in Table \ref{C_confidence}. As can be seen, regardless of the threshold value, our method consistently outperforms NN confidence.

\begin{table}[h]
\centering
\begin{tabular}{clcc}
\toprule
\textbf{Retain Threshold} & \multicolumn{1}{c}{\textbf{Method}} & \multicolumn{1}{c}{\textbf{Recall}} & \multicolumn{1}{c}{\textbf{Inference Accuracy}} \\
\midrule
\multirow{2}{*}{60\%} & NN Confidence & 93.18\scriptsize{$\pm$2.34} & 77.2\scriptsize{$\pm$5.5} \\
 & \textbf{ABL-Refl} ($C=0.6$) & \textbf{99.31\scriptsize{$\pm$0.84}} & \textbf{95.8\scriptsize{$\pm$2.8}} \\
\midrule
\multirow{2}{*}{70\%} & NN Confidence & 88.60\scriptsize{$\pm$2.66} & 70.1\scriptsize{$\pm$5.7} \\
 & \textbf{ABL-Refl} ($C=0.7$) & \textbf{99.25\scriptsize{$\pm$0.84}} & \textbf{94.5\scriptsize{$\pm$2.9}} \\
\midrule
\multirow{2}{*}{80\%} & NN Confidence & 82.64\scriptsize{$\pm$2.78} & 64.3\scriptsize{$\pm$6.2} \\
 & \textbf{ABL-Refl} ($C=0.8$) & \textbf{99.04\scriptsize{$\pm$0.85}} & \textbf{93.5\scriptsize{$\pm$3.2}} \\
\midrule
\multirow{2}{*}{90\%} & NN Confidence & 71.05\scriptsize{$\pm$3.01} & 52.1\scriptsize{$\pm$6.2} \\
 & \textbf{ABL-Refl} ($C=0.9$) & \textbf{98.86\scriptsize{$\pm$0.89}} & \textbf{91.2\scriptsize{$\pm$3.5}} \\
\bottomrule
\end{tabular}
\caption{Recall and inference accuracy for different thresholds of intuitive output retained (In ABL-Refl, the threshold is controlled by $C$ as a soft margin and not a strict boundary).}
\label{C_confidence}
\end{table}

\section{Additional Experiment on Solving Combinatorial Optimization Problems on Graphs}
\label{app_add}

In this section, we present an additional experiment on solving combinatorial optimization problems on graphs, finding the maximum independent set. In this experiment, we will demonstrate how our method can easily extend across varied reasoning scenarios. 

\paragraph{Dataset and Settings.} 

The input is the same as in Section \ref{exp3} for solving the maximum clique, given a graph $G=(V,E)$ with $|V|=n$ nodes, but in this section, we aim for the output $\boldsymbol{y}\in\{0,1\}^n$ where the set of value 1 collectively constitutes the maximum independent set. While the two problems share similarities, they exhibit distinct reasoning capabilities: cliques rely on high homophily, whereas an independent set demonstrates significant heterophily. Generally, it is challenging for graph neural networks to simultaneously handle both scenarios effectively.

We utilize the same structure of graph neural networks as in Section \ref{exp3}. For the reasoning part, we continue to use Gurobi as the symbolic solver, and $\mathcal{KB}$ remains the basic mathematical definition of an independent set, i.e., no two nodes are connected by an edge. For consistency measurement, we adopt a similar definition in Section \ref{exp3} as follows: one point is awarded for each pair of vertices if they are not connected by an edge; additionally, if the output set is indeed an independent set, the size of the output set multiplied by 10 is added. We may see that although the nature of the reasoning becomes entirely opposite compared to solving the maximum clique, we are able to flexibly transition to the new scenario with minimal changes. 

\paragraph{Results.}

We report the results in Table \ref{result_max_ind_set}. We may see that our method significantly outperforms compared methods. Additionally, when compared to the results in Table \ref{C_clique}, it can be observed that the performance of other baselines has declined when switching from finding maximum cliques to this task of finding maximum independent set. However, the performance of ABL-Refl has remained near perfect.

\begin{table}[h]
    \centering
    \begin{tabular}{llcccc}
    \toprule
     \multicolumn{2}{c}{\multirow{2}{*}{\textbf{Method}}} & \multicolumn{4}{c}{\textbf{Dataset}} \\
    \cmidrule{3-6}
    & & ENZYMES & PROTEINS & IMDB-Binary & COLLAB \\
    \midrule
    \multicolumn{2}{l}{Erdos} & 0.821\scriptsize{$\pm$0.125} & 0.903\scriptsize{$\pm$0.114} & 0.515\scriptsize{$\pm$0.310} & 0.886\scriptsize{$\pm$0.198} \\
    \multicolumn{2}{l}{Neural SFE} & 0.775\scriptsize{$\pm$0.155} & 0.729\scriptsize{$\pm$0.205} & 0.679\scriptsize{$\pm$0.287} & 0.392\scriptsize{$\pm$0.253} \\
    \midrule
    \multirow{3}{*}{\textbf{ABL-Refl}} & $C=0.7$ & \textbf{0.989\scriptsize{$\pm$0.022}} & \textbf{0.958\scriptsize{$\pm$0.029}} & \textbf{0.964\scriptsize{$\pm$0.026}} & \textbf{0.987\scriptsize{$\pm$0.016}} \\
    & $C=0.8$ & \textbf{0.986\scriptsize{$\pm$0.026}} & \textbf{0.954\scriptsize{$\pm$0.053}} & \textbf{0.960\scriptsize{$\pm$0.037}} & \textbf{0.985\scriptsize{$\pm$0.016}} \\
    & $C=0.9$ & \textbf{0.980\scriptsize{$\pm$0.025}} & \textbf{0.942\scriptsize{$\pm$0.051}} & \textbf{0.952\scriptsize{$\pm$0.021}} & \textbf{0.975\scriptsize{$\pm$0.021}} \\
    \bottomrule
    \end{tabular}
    \caption{Approximation ratios on finding maximum maximum independent set.}
    \label{result_max_ind_set}
\end{table}

\end{document}